\documentclass[lettersize,journal]{IEEEtran}
\usepackage{amsfonts,bm,booktabs,cite,graphicx,hyperref,multirow,textcomp,url}
\usepackage{amsmath}
\usepackage{algorithmic}
\usepackage{algorithm}
\usepackage{array}
\usepackage{amssymb}
\usepackage[caption=false,font=normalsize,labelfont=sf,textfont=sf]{subfig}
\usepackage{textcomp}
\usepackage{stfloats}
\usepackage{url}
\usepackage{verbatim}
\usepackage{graphicx}
\usepackage{cite}
\usepackage{bbding}
\usepackage[flushleft]{threeparttable}
\begin{document}

\title{A Spatiotemporal Semantic Importance-Guided Unified Compression and Editing Framework for AI-Generated Videos}

\author{Xihua Sheng,~\IEEEmembership{Member,~IEEE}, 
Dong Liu,~\IEEEmembership{Senior Member,~IEEE}, 
Chang Wen Chen,~\IEEEmembership{Life Fellow,~IEEE}\\
\thanks{Date of current version \today.}
\thanks{X. Sheng and C. Chen are with the Department of Computing, The Hong Kong Polytechnic University, Hong Kong SAR, China (e-mail: xihua.sheng@polyu.edu.hk, changwen.chen@polyu.edu.hk).\par
D. Liu is with the MOE Key Laboratory of Brain-Inspired Intelligent
Perception and Cognition, University of Science and Technology of China,
Hefei 230093, China (e-mail: dongeliu@ustc.edu.cn).
}

\thanks{Corresponding Author: Chang Wen Chen.}
}
\markboth{Journal of \LaTeX\ Class Files,~Vol.~14, No.~8, August~2021}%
{Shell \MakeLowercase{\textit{et al.}}: A Sample Article Using IEEEtran.cls for IEEE Journals}



\maketitle

\begin{abstract}
AI-generated videos are rapidly increasing in volume, duration, and resolution, creating growing demands for efficient storage and transmission. Unlike natural videos captured from the physical world, AI-generated videos are samples from a learned generative distribution, where semantic structures are critical to content consistency, while many local textures and stochastic details can be plausibly regenerated. This distinction suggests that compression should preserve semantically important spatiotemporal information rather than reconstruct every pixel of a particular generative sample. Beyond reconstruction, AI-generated videos also create a practical need for prompt-based editing, where users expect to modify generated content while preserving its original spatiotemporal semantics. Motivated by these observations, we propose a unified compression and editing framework for AI-generated videos that incorporates a frozen video generator as a reusable generative prior. Within this framework, we design three spatiotemporal semantic importance-guided techniques that respectively address what to transmit, how much to transmit, and how to use the transmitted side information. First, an innovation selection method projects the latent discrepancy using spatiotemporal semantic importance, so that the selected innovations prioritize semantic invariants over replaceable generative variations. Second, a frame-adaptive bit allocation method estimates the nonuniform semantic demands of latent frames and allocates more innovations to frames requiring stronger semantic preservation. Third, a unified reconstruction and editing method continuously adjusts the influence of the transmitted side information, enabling the same compressed representation to provide strong guidance for faithful reconstruction or serve as a flexible semantic anchor for structure-preserving prompt-driven editing. Extensive experiments on videos generated by multiple video generation models show that our codec achieves superior rate-perception performance over traditional, neural, and generative video codecs, while enabling structure-preserving prompt-based editing from the same bitstream.

\end{abstract}


\section{Introduction}
Artificial Intelligence (AI) video generation is emerging as a new paradigm for content creation. It is becoming increasingly popular across a wide range of applications such as short-form video creation and advertising. As AI-generated videos rapidly grow in volume, duration, and resolution, they are creating substantial storage and transmission demands. This raises a direct but underexplored problem: \emph{how should AI-generated videos be compressed?}

Existing video coding methods~\cite{wiegand2003overview,sullivan2012overview,bross2021developments} mostly assume that videos are natural signals captured by cameras. A natural video records the physical world: object boundaries, textures, motion trajectories, and illumination changes all come from a real scene, and the goal of compression is to preserve this external signal under a bitrate constraint. AI-generated videos are fundamentally different. They are not measurements of a real scene, but samples drawn from the distribution learned by a generative model~\cite{xing2024survey}. In such videos, identity, layout, object relations, and motion trends usually define the spatiotemporal semantics of the content, while many local textures, small appearance changes, and stochastic details are only one possible realization of the generation process. In other words, not all pixel-level differences in an AI-generated video are equally meaningful: some differences break identity, silhouette, or motion consistency, while others can be regenerated differently and still remain plausible. Directly compressing AI-generated videos as natural videos therefore ignores the statistical origin and semantic structure of generated content. This distinction changes the coding objective. For AI-generated videos, the goal should not be to reconstruct every pixel, but to preserve the information that determines semantic consistency. 

Beyond reconstruction, this semantic view also changes how a compressed bitstream can be used at the receiver side. In conventional video coding, the bitstream is primarily designed for decoded playback. When further modifications are desired, such as transferring the style, changing object attributes, removing the watermarks of the generated video while preserving its original structure and motion, a separate editing model is commonly applied to the reconstructed video. Such a  model needs to analyze the video again, for example by encoding it into a latent space to extract the structural and motion information required for prompt-based editing~\cite{ceylan2023pix2video,chai2023stablevideo,mai2026easyv2v,litman2026editctrl,feng2024ccedit}. However, this spatiotemporal semantic information has already been conveyed by the compressed bitstream. Re-extracting it from the reconstructed video therefore introduces redundant processing and additional computational complexity.
In this setting, the compressed bitstream should not merely describe a one-time reconstruction. It should also carry reusable semantic constraints that support controlled editing.

Video generation models~\cite{wan2025wan,lin2024open,blattmann2023stable} provide a new opportunity to realize these goals. A video generator~\cite{ho2020denoising,song2020denoising,lipman2022flow} has learned to transform noisy latent variables into video content with plausible appearance, structure, and motion. When a frozen video generator is integrated into a codec, it can serve as a reusable video prior: the generator restores replaceable details, while the bitstream only needs to transmit compact side information that constrains the generation process toward the source video. In this sense, compression no longer has to independently describe the complete video signal. Meanwhile, because the generator is prompt-conditioned, the same side information can be combined with a user-defined prompt to produce an edited video that follows the editing intention while preserving the source spatiotemporal semantics. This forms a unified AI-generated video coding framework, where one compressed representation supports both high-quality reconstruction and prompt-based editing, without training a new compression model or introducing a separate post-reconstruction editing model.

However, turning this opportunity into a practical codec faces three key problems. First, side-information selection needs to account for spatiotemporal semantic importance. If information is selected only according to latent errors, the bitstream may spend bits on replaceable stochastic textures instead of protecting the semantic structures that determine content consistency. Second, the side-information budget needs to reflect the nonuniform spatiotemporal semantic importance of latent frames. Uniform allocation may waste bits on semantically stable frames with substantial temporal redundancy, while providing insufficient information for frames that require stronger semantic preservation. Third, reconstruction and editing impose different requirements on the same side information. Reconstruction needs strong constraints to recover the source video, while editing needs sufficient flexibility to follow a new prompt. Overly rigid constraints suppress editing, whereas overly weak constraints cause structural drift. Therefore, the codec needs a controllable mechanism for using the same bitstream under different reconstruction and editing requirements.

To address these problems, we propose a spatiotemporal semantic importance-guided unified compression and editing framework for AI-generated videos. First, we propose a spatiotemporal semantic importance-guided innovation selection method to determine what information should be transmitted. It projects the latent discrepancy according to spatiotemporal semantic importance and selects innovations that prioritize semantic invariants over replaceable generative variations. Second, we propose a spatiotemporal semantic importance-guided bit allocation method to determine how much information should be assigned to each latent frame. It evaluates frame-level semantic demand and allocates more innovations to frames requiring stronger semantic preservation. Third, we propose a spatiotemporal semantic importance-constrained unified reconstruction and editing method to determine how the transmitted side information should be used at the receiver. It continuously adjusts the influence of the bitstream constraint, allowing the same compressed representation to provide strong guidance for source reconstruction or serve as a flexible semantic anchor for prompt-driven editing.

To summarize, our main contributions are threefold:
\begin{itemize}
    \item We propose, to the best of our knowledge, the first unified compression and editing framework specifically designed for AI-generated videos, where one compact bitstream supports both high-quality reconstruction and structure-preserving prompt-based editing.
    \item We propose three spatiotemporal semantic importance-guided techniques: an innovation selection method that prioritizes semantically critical latent information, a frame-adaptive bit allocation method that adapts the innovation budget to frame-level semantic demand, and a unified reconstruction and editing method that balances transmitted spatiotemporal semantics with prompt-induced editing semantics.

    \item Extensive experiments on AI-generated videos from multiple generation models demonstrate that our framework achieves superior rate-perception performance against traditional, neural, and generative video codecs, while further enabling structure-preserving prompt-based editing from the same compressed bitstream.
\end{itemize}

The remainder of this paper is organized as follows. Section~\ref{related_work} reviews related work. Section~\ref{sec:method} describes our proposed methods. Section~\ref{sec:experiments} presents experimental results.  Section~\ref{sec:conclusion} concludes the paper and
discusses the future work.

\section{Related Work}\label{related_work}
\subsection{Generative Video Compression}
Generative video compression reconstructs perceptually realistic videos at low bitrates by incorporating generative priors into the codec. Existing work can be grouped into generative adversarial network (GAN)~\cite{goodfellow2020generative}-based, vector-quantization-based, visual autoregressive-based, diffusion-based, and training-free methods.

GAN-based methods introduce adversarial learning into learned video compression to improve perceptual realism at low bitrates. Mentzer et al.~\cite{mentzer2022neural} proposed the first GAN-based neural video compression system, where detail synthesis is conditioned on the warped previous reconstruction and then propagated through optical flow, showing that adversarial learning is crucial for subjective visual quality. Yang et al.~\cite{yang2022perceptual} proposed a recurrent conditional GAN-based video codec, in which the discriminator is conditioned on spatial-temporal features, encouraging reconstructions to be both photo-realistic and temporally coherent. Li et al.~\cite{li2023high} proposed to improve visual fidelity by enhancing newly emerged regions with confidence-based feature reconstruction and suppressing checkerboard artifacts with a periodic compensation loss. Du et al.~\cite{du2024cgvc} proposed to combine contextual coding, transformer-convolution autoencoders, entropy modeling, and GAN-based perceptual enhancement for generative video compression. Liang et al.~\cite{liang2026perceptual} proposed a rank-chain GAN loss function to improve perceptual quality across variable bitrates. These methods demonstrate that adversarial learning can effectively improve perceptual quality.

Vector-quantization-based methods shift video compression from the pixel domain to a generative latent or token space. Qi et al.~\cite{qi2025generative} proposed a method that performs image and video compression in the latent space of a generative VQ-VAE, showing that such a space is sparser, more semantic, and better aligned with perceptual quality than pixel space at ultra-low bitrates. Guo et al.~ \cite{guo2025generative} further explored continuous generative tokenization for perceptual video compression, where a pretrained tokenizer projects video frames into a perceptually aligned latent domain and the codec is redesigned around this representation. Zheng et al.~\cite{zheng2026generative} proposed to represent videos with compact one-dimensional latent tokens, allowing the codec to reduce spatial redundancy and model long-term temporal context through a lightweight memory mechanism. Dang et al.~\cite{dang2026gvc} pursued real-time generative video compression by designing a token-based codec with efficient generative reconstruction. These methods show that generative token spaces can improve perceptual compression by representing videos with more compact and semantically meaningful units. 

Visual autoregressive-based methods formulate generative video compression as coarse-to-fine token modeling. Li et al.~\cite{li2026progvc} proposed to encode videos into hierarchical residual token maps, where coarse tokens are transmitted for progressive reconstruction and fine-scale tokens can be predicted by an autoregressive context model at the decoder. This design links progressive transmission, entropy modeling, and generative detail synthesis within a token prediction framework.

Diffusion-based methods exploit diffusion or flow-based priors to synthesize perceptually plausible details at low bitrates. Mao et al.~\cite{mao2026generative} used a video diffusion prior for sequence-level generative refinement, improving temporal stability over frame-wise generative enhancement. Gao et al.~\cite{gao2025givic} incorporated diffusion sampling into implicit neural representations for coarse-to-fine spatiotemporal reconstruction. Zhang et al.~\cite{zhang2026lgvf} designed a spatio-temporal diffusion framework to reduce error accumulation and handle semantic mutation during low-bitrate reconstruction. Xue et al.~\cite{xue2026single} developed a single-step diffusion codec with semantic-temporal guidance to improve reconstruction efficiency and temporal coherence. Ma et al.~\cite{ma2025diffusion,ma2025diffvc,ma2026diffvc} studied real-time and one-step diffusion-based perceptual video compression, reducing the sampling cost of diffusion decoders while maintaining perceptual quality. Li et al.~\cite{li2026yoda} embedded temporal reference information into a one-step diffusion compressor to better exploit spatiotemporal correlations. Other methods improve controllability by transmitting compact semantic or structural cues. Wang et al.~\cite{wang2026low} used textual, appearance, and motion cues to condition diffusion reconstruction, and Wang et al.~\cite{wang2026t} introduced sparse semantic motion trajectories to guide generative video coding. These methods show the strong potential of diffusion priors for perceptual video compression, especially at ultra-low bitrates.

Training-free methods reuse pretrained generative priors for video compression without updating model parameters. Ling et al.~\cite{ling2026free} used a video diffusion model to formulate video coding as latent trajectory compression, where reverse-channel coding transmits stochastic choices and adaptive inter-group mechanisms improve temporal coherence. Zeng et al.~\cite{zeng2026gvcc} transformed a deterministic rectified-flow sampler into a stochastic coding process, allowing compact codebook-selected innovations to steer a frozen video generator toward the source video. Gao et al.~\cite{gao2026zerogvc} used pretrained autoregressive diffusion priors and represented predictive frames through compact combinations of reproducible codebook vectors. These methods show that frozen generative models can be used as powerful decoders and can be steered by compact side information without codec training.

Overall, existing generative video compression methods have substantially advanced low-bitrate video reconstruction by introducing powerful generative priors. However, they are mainly developed for natural videos, where the source is treated as an observed signal to be approximated under a bitrate constraint. Even with advanced video generation models, the selection and allocation of transmitted information are often guided by latent errors of the observed signal. This differs from AI-generated videos, where the source itself is a sample from a learned generative distribution and many local stochastic details are replaceable rather than essential. Therefore, AI-generated video compression requires a semantic-aware coding principle that prioritizes spatiotemporal information critical to content consistency.

\subsection{AI-Generated Video Compression}
Recently, AI-generated content has begun to receive attention in image compression. Duan et al.~\cite{duan2024pku} studied neural compression for AI foundation model generated images and constructed PKU-AIGI-500K, a large-scale benchmark containing diverse prompts and images generated by multiple foundation models. Their benchmark shows that conventional codecs and learned codecs trained on natural images do not necessarily generalize well to AI-generated images, revealing a clear domain gap in generated-content compression. Duan et al.~\cite{duan2024neural} further extended this benchmark study by analyzing the characteristics of AI-generated images and proposing a cross-attention transformer codec that exploits the relationship between generated images and their text prompts. Chen et al.~\cite{chen2024stable} explored a different direction by using Stable Diffusion as a cross-modal decoder for layered AI-generated image compression. Their method represents an image with semantic, structural, and texture layers, showing that a generative model can act as a decoder when compact multimodal priors are transmitted.

These works provide important evidence that AI-generated content deserves specialized compression designs. However, compression specifically designed for AI-generated videos remains largely unexplored. To the best of our knowledge, our work is the first unified compression and editing framework for AI-generated videos.

\subsection{Generative Video Editing}
Generative video editing aims to modify existing videos according to text instructions or user-specified conditions while preserving source structure and temporal coherence. Early diffusion-based methods often adapt pretrained text-to-image or image diffusion models to videos. Ceylan et al.~\cite{ceylan2023pix2video} proposed a training-free method that edits an anchor frame and propagates the edit through self-attention feature injection, showing that image diffusion priors can be extended to text-guided video editing. Zhang et al.~\cite{zhang2023towards} studied consistent video editing with text-to-image diffusion models and showed that newly introduced temporal modules may cause feature-space covariate shift, leading to temporal inconsistency and semantic disparity. Chai et al.~\cite{chai2023stablevideo} introduced temporal dependency into diffusion editing by propagating appearance information across frames with layered representations, improving consistency for edited objects. Feng et al.~\cite{feng2024ccedit} separated structure and appearance control with multiple branches, enabling diverse edits such as style transfer, foreground modification, and background replacement. Mai et al.~\cite{mai2026easyv2v} developed an instruction-based video editing framework with flexible text, mask, reference, and temporal controls. Litman et al.~\cite{litman2026editctrl} improved editing efficiency by focusing computation on local masked regions while using a lightweight global context module to maintain temporal consistency. Munir et al.~\cite{munir2026smoothdiffusion} proposed a real-time streaming editing method with adaptive feature caching and motion-guided attention to reduce redundant computation while preserving temporal coherence.

These methods demonstrate impressive progress in controllable and temporally consistent video editing. A common property is that, given an input video, they need to analyze the video and construct conditions for prompt-guided generation. Depending on the method, such conditions may come from VAE latents, motion, or structural representations. Interestingly, these conditions are closely related to the information that a generative codec needs to transmit for preserving source content. This connection suggests that video compression and video editing need not be treated as two isolated stages. In our framework, the side information transmitted for compression is selected and allocated according to spatiotemporal semantic importance, and can therefore be reused as a controllable semantic condition for prompt-based editing.
\begin{figure*}[t]
  \centering
  \includegraphics[width=\linewidth]{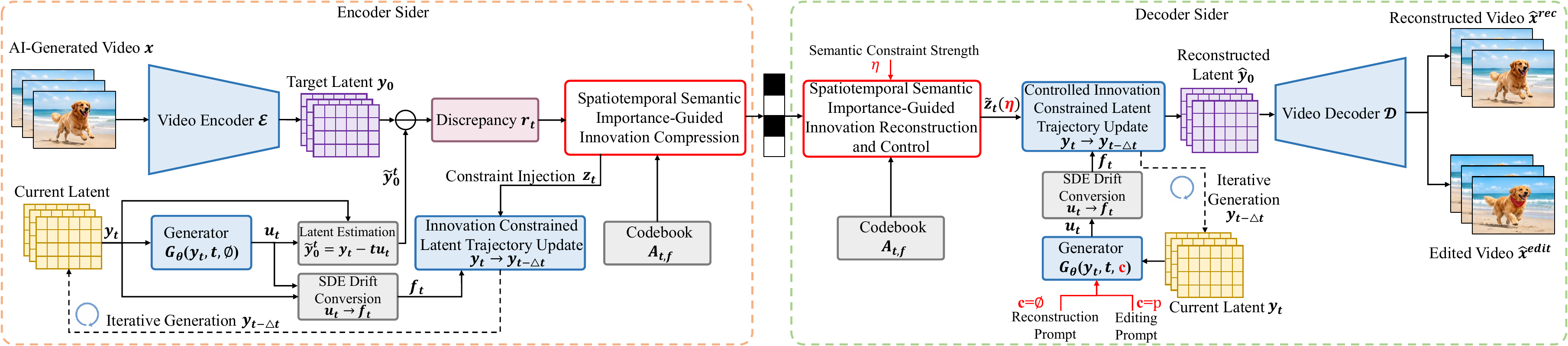}
  \caption{Overview of the proposed spatiotemporal semantic importance-guided unified compression and editing framework for AI-generated videos.}
  \label{fig:framework}
\end{figure*}
\section{Method}\label{sec:method}
\subsection{Framework Overview}
\label{sec:framework_overview}

We first present an overview of the proposed framework, as illustrated in Fig.~\ref{fig:framework}.

\subsubsection{Frozen Generative Video Representation}
Given an input AI-generated video $\mathbf{x}$, a frozen video autoencoder $\mathcal{E}$ maps it into the latent space of a pretrained video generator:
\begin{equation}
    \mathbf{y}_0
    =
    \mathcal{E}(\mathbf{x}),
    \label{eq:target_latent}
\end{equation}
where $\mathbf{y}_0\in \mathbb{R}^{C\times F_{\ell}\times H_{\ell}\times W_{\ell}}$, $C$, $F_{\ell}$, $H_{\ell}$, and $W_{\ell}$ denote the latent channel, temporal, height, and width dimensions, respectively. The $f$-th latent frame is denoted by $\mathbf{y}_{0,f}$.

The latent variable $\mathbf{y}_0$ serves as the coding target. Starting from an initial noisy latent $\mathbf{y}_T$, a frozen rectified-flow video generator $G_\theta$~\cite{liu2022flow} predicts the velocity at each generation step:
\begin{equation}
    \mathbf{u}_t
    =
    G_\theta
    \left(
        \mathbf{y}_t,
        t,
        \varnothing
    \right),
    \label{eq:generator_velocity}
\end{equation}
where $\theta$ denotes the frozen generator parameters and $\varnothing$ denotes the null-text condition used during encoding. Under the linear rectified-flow interpolation, the predicted velocity provides an estimate of the clean video latent:
\begin{equation}
    \widetilde{\mathbf{y}}_0^{\,t}
    =
    \mathbf{y}_t
    -
    t\mathbf{u}_t.
    \label{eq:clean_latent_estimation}
\end{equation}
The latent discrepancy at generation step $t$ is then defined as
\begin{equation}
    \mathbf{r}_t
    =
    \mathbf{y}_0
    -
    \widetilde{\mathbf{y}}_0^{\,t}.
    \label{eq:latent_discrepancy}
\end{equation}
This discrepancy specifies the video-dependent information that has not yet been recovered by the shared generative prior.

\subsubsection{Codebook-Based Stochastic Innovation Coding}
To represent the information remaining in $\mathbf{r}_t$, we adopt the codebook-based stochastic flow coding formulation in~\cite{ohayon2025compressed,vaisman2026turbo,zeng2026gvcc}. This formulation converts the deterministic rectified-flow trajectory into a stochastic process whose innovations can be represented using a reproducible random codebook.

At each generation step $t$ and latent frame $f$, the encoder and decoder use a shared seed $\omega$ to construct the same candidate codebook:
\begin{equation}
    \mathbf{a}_{t,f,k}
    =
    \mathcal{R}
    \left(
        \omega,
        t,
        f,
        k
    \right),
    \label{eq:codebook_atom}
\end{equation}
\begin{equation}
    \mathcal{A}_{t,f}
    =
    \left\{
        \mathbf{a}_{t,f,k}
    \right\}_{k=1}^{K},
    \label{eq:innovation_codebook}
\end{equation}
where $\mathcal{R}$ is a deterministic Gaussian sampler and $K$ is the number of candidate atoms. Because the codebook can be reproduced from the shared seed, its elements do not need to be transmitted.

The latent discrepancy $\mathbf{r}_{t,f}$ provides the target for selecting atoms from $\mathcal{A}_{t,f}$. Instead of directly transmitting $\mathbf{r}_{t,f}$, the encoder selects a small number of signed atoms that provide a sparse codebook-based representation of the required trajectory correction. Let $i_{t,f,m}$ and $s_{t,f,m}\in\{-1,+1\}$ denote the index and sign of the $m$-th selected atom, respectively. The frame-wise innovation is constructed as
\begin{equation}
    \mathbf{z}_{t,f}
    =
    \frac{1}{\nu_{t,f}}
    \sum_{m=1}^{M_{t,f}}
    s_{t,f,m}\,
    \mathbf{a}_{t,f,i_{t,f,m}},
    \label{eq:framewise_innovation}
\end{equation}
where $M_{t,f}$ is the number of selected atoms and $\nu_{t,f}$ normalizes the signed sum to unit variance. The frame-wise innovations are stacked along the temporal dimension to form $\mathbf{z}_t$.

To incorporate these innovations into the rectified-flow trajectory, the generator-predicted velocity is converted into an SDE drift~\cite{ohayon2025compressed,vaisman2026turbo,zeng2026gvcc}. For the linear interpolation used by the frozen generator, the score is obtained from $\mathbf{u}_t$ as
\begin{equation}
    \nabla_{\mathbf{y}_t}
    \log p_t(\mathbf{y}_t)
    =
    -
    \frac{
        \mathbf{y}_t
        +
        (1-t)\mathbf{u}_t
    }{
        t
    }.
    \label{eq:velocity_to_score}
\end{equation}
The corresponding SDE drift is
\begin{equation}
    \mathbf{f}_t
    =
    \mathbf{u}_t
    -
    \frac{g_t^2}{2}
    \nabla_{\mathbf{y}_t}
    \log p_t(\mathbf{y}_t),
    \label{eq:sde_drift}
\end{equation}
where $g_t$ controls the stochastic strength. The latent trajectory is updated using
\begin{equation}
    \mathbf{y}_{t-\Delta t}
    =
    \mathbf{y}_t
    -
    \mathbf{f}_t\Delta t
    +
    g_t\sqrt{\Delta t}\,
    \mathbf{z}_t.
    \label{eq:stochastic_latent_update}
\end{equation}

The selected innovation steers the trajectory toward the target latent $\mathbf{y}_0$. After each update, the generator produces a new clean-latent estimate, from which the encoder calculates an updated discrepancy and selects the innovations for the next generation step. At the decoder, the transmitted indices and signs reproduce the same innovations, allowing the same stochastic trajectory to be replayed without access to the source video.

\subsubsection{Spatiotemporal Semantic Importance-Guided Innovation Selection}

The latent discrepancy contains both semantic information that should remain consistent with the source video and local generative variations that can be plausibly regenerated. Treating all latent positions equally may therefore spend coding capacity on replaceable details. We propose a spatiotemporal semantic importance-guided innovation selection method to determine what information should be transmitted. For each generation step and latent frame, the method constructs a position-wise importance map $\mathbf{W}_{t,f}$ and uses it to weight the latent discrepancy. Candidate atoms are then evaluated according to their ability to represent the semantically important components of the discrepancy while maintaining local temporal consistency. More details are presented in Section~\ref{sec:semantic_projection}.

\subsubsection{Spatiotemporal Semantic Importance-Guided Bit Allocation}
The semantic coding demand is nonuniform across latent frames. Frames containing prominent structures, temporal changes, or substantial unrecovered information generally require stronger trajectory constraints than semantically stable frames. We propose a spatiotemporal semantic importance-guided bit allocation method to determine how much information should be transmitted for each latent frame. The method calculates a frame-level coding demand $L_{t,f}$, converts it into an allocation score $\rho_{t,f}$, and determines the corresponding innovation budget $M_{t,f}$. 
Thus, the innovation-selection method determines which atoms are transmitted under the assigned budget, while the bit-allocation method determines the size of that budget. More details are presented in Section~\ref{sec:temporal_allocation}.

\subsubsection{Spatiotemporal Semantic Importance-Constrained Unified Reconstruction and Editing}
Given the transmitted bitstream, the decoder can operate in either source-reconstruction or prompt-based editing mode. We propose a spatiotemporal semantic importance-constrained unified reconstruction and editing method that treats the recovered innovations as controllable semantic constraints.
For source reconstruction, the recovered innovations are fully applied under the null-text condition. For prompt-based editing, their contribution is relaxed while a user-defined prompt guides the generation process. The same bitstream can therefore support faithful reconstruction and structure-preserving editing without re-encoding the source video. Details are presented in Section~\ref{sec:prompt_editing}.

\subsection{Spatiotemporal Semantic Importance-Guided Innovation Selection}
\label{sec:semantic_projection}

Given the latent discrepancy $\mathbf{r}_{t,f}$ and candidate codebook $\mathcal{A}_{t,f}$, the encoder needs to select $M_{t,f}$ correction directions. Existing codebook-based methods~\cite{zeng2026gvcc} commonly evaluate each candidate according to its correlation with the unweighted latent discrepancy:
\begin{equation}
    B_{t,f,k}^{\mathrm{lat}}
    =
    \left|
        \left\langle
            \mathbf{a}_{t,f,k},
            \mathbf{r}_{t,f}
        \right\rangle
    \right|,
    \label{eq:latent_matching}
\end{equation}
where $\langle\mathbf{a},\mathbf{b}\rangle$ denotes the inner product between two latent tensors. This criterion treats all latent positions as equally important. However, discrepancies associated with persistent structures and temporal evolution are more important to source consistency than replaceable stochastic details.

\begin{figure}[t]
  \centering
  \includegraphics[width=\linewidth]{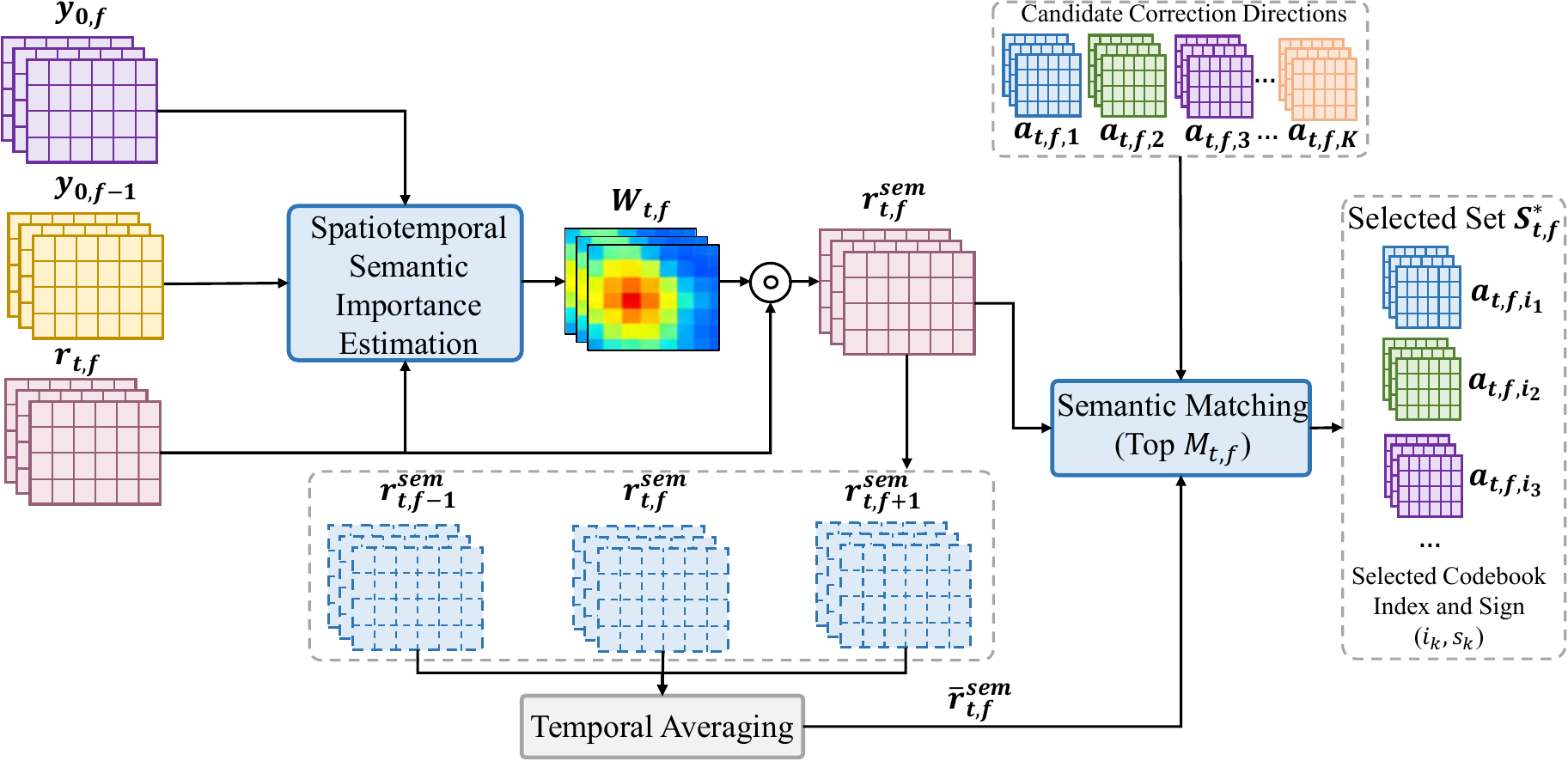}
\caption{Illustration of the spatiotemporal semantic importance-guided innovation selection method. Here, $\mathbf{y}_{0,f}$ and $\mathbf{y}_{0,f-1}$ denote the current and preceding source latent frames, respectively. $\mathbf{r}_{t,f}$ denotes the unrecovered latent discrepancy for latent frame $f$ at generation step $t$. $\mathbf{W}_{t,f}$ is the position-wise spatiotemporal semantic importance map. $\mathbf{r}_{t,f}^{\mathrm{sem}}$ is the importance-weighted discrepancy and $\overline{\mathbf{r}}_{t,f}^{\mathrm{sem}}$ is the temporally averaged discrepancy.}
  \label{fig:semantic_selection}
\end{figure}

As illustrated in Fig.~\ref{fig:semantic_selection}, our method evaluates candidate atoms in a spatiotemporal semantic importance-weighted latent space. We construct a position-wise spatiotemporal importance map from the spatial variation of the source latent, the temporal variation between adjacent latent frames, and the discrepancy remaining at the current generation step:
\begin{equation}
\begin{aligned}
    \mathbf{W}_{t,f}
    ={}&
    \mathbf{1}
    +
    \lambda_s
    \mathcal{N}
    \left(
        \left|
            \nabla_h\mathbf{y}_{0,f}
        \right|
        +
        \left|
            \nabla_w\mathbf{y}_{0,f}
        \right|
    \right)
    \\
    &+
    \lambda_t
    \mathcal{N}
    \left(
        \left|
            \mathbf{y}_{0,f}
            -
            \mathbf{y}_{0,f-1}
        \right|
    \right)
    +
    \lambda_r
    \mathcal{N}
    \left(
        \left|
            \mathbf{r}_{t,f}
        \right|
    \right).
    \label{eq:semantic_importance}
\end{aligned}
\end{equation}
The spatial-gradient term identifies latent positions associated with structural variation. The temporal-difference term identifies changes in video content across adjacent latent frames. The discrepancy term identifies information that remains unrecovered at the current generation step.
The operator $\mathcal{N}(\cdot)$ normalizes the three cues to comparable scales, while $\lambda_s$, $\lambda_t$, and $\lambda_r$ control their relative contributions. Unlike the frame-level quantity introduced in Section~\ref{sec:temporal_allocation}, $\mathbf{W}_{t,f}$ retains the spatial and channel dimensions of the latent frame and therefore provides position-wise importance weights.
The importance-weighted discrepancy is defined as
\begin{equation}
    \mathbf{r}_{t,f}^{\mathrm{sem}}
    =
    \mathbf{W}_{t,f}
    \odot
    \mathbf{r}_{t,f},
    \label{eq:semantic_projected_residual}
\end{equation}
where $\odot$ denotes element-wise multiplication. This operation increases the contribution of discrepancies at important latent positions during codebook matching without modifying the transmitted codebook itself.

Selecting atoms independently for each latent frame may produce temporally inconsistent corrections. We therefore construct a local temporal reference from the importance-weighted discrepancies of neighboring latent frames:
\begin{equation}
    \overline{\mathbf{r}}_{t,f}^{\mathrm{sem}}
    =
    \frac{1}{|\Omega_f|}
    \sum_{j\in\Omega_f}
    \mathbf{r}_{t,j}^{\mathrm{sem}},
    \label{eq:temporal_semantic_reference}
\end{equation}
where $\Omega_f$ denotes the set of valid temporal neighbors of latent frame $f$. This reference introduces local temporal context into candidate evaluation without requiring additional side information.

The semantic matching score for candidate atom $\mathbf{a}_{t,f,k}$ is
\begin{equation}
    B_{t,f,k}^{\mathrm{sem}}
    =
    \left|
        \left\langle
            \mathbf{a}_{t,f,k},
            \mathbf{r}_{t,f}^{\mathrm{sem}}
        \right\rangle
    \right|
    +
    \lambda_T
    \left|
        \left\langle
            \mathbf{a}_{t,f,k},
            \overline{\mathbf{r}}_{t,f}^{\mathrm{sem}}
        \right\rangle
    \right|,
    \label{eq:semantic_matching_benefit}
\end{equation}
where $\lambda_T$ controls the contribution of the neighboring-frame reference. Given the allocated budget $M_{t,f}$, the selected subset is
\begin{equation}
    \mathcal{S}_{t,f}^{*}
    =
    \underset{
        \substack{
            \mathcal{S}\subseteq\{1,\ldots,K\},\\
            |\mathcal{S}|=M_{t,f}
        }
    }{\arg\max}
    \sum_{k\in\mathcal{S}}
    B_{t,f,k}^{\mathrm{sem}}.
    \label{eq:semantic_selection}
\end{equation}
For each selected atom, its sign is determined from its correlation with the current importance-weighted discrepancy:
\begin{equation}
    s_{t,f,m}
    =
    \operatorname{sign}
    \left(
        \left\langle
            \mathbf{a}_{t,f,i_{t,f,m}},
            \mathbf{r}_{t,f}^{\mathrm{sem}}
        \right\rangle
    \right).
    \label{eq:semantic_atom_sign}
\end{equation}
The selected indices and signs are written into the bitstream. Because the decoder constructs the same candidate codebook from the shared seed, it can reproduce the selected signed atoms without receiving either $\mathbf{W}_{t,f}$ or $\mathbf{r}_{t,f}^{\mathrm{sem}}$.

\begin{figure}[t]
  \centering
  \includegraphics[width=\linewidth]{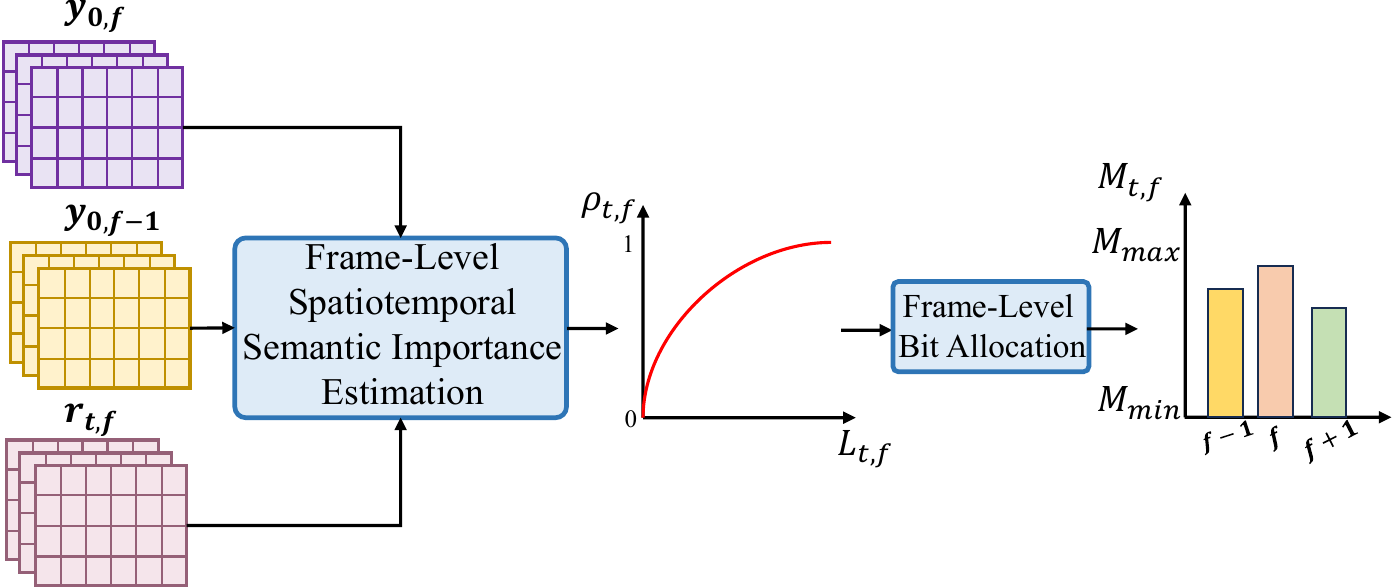}
  \caption{Illustration of the proposed spatiotemporal semantic importance-guided bit allocation method. Here, $\mathbf{y}_{0,f}$ and $\mathbf{y}_{0,f-1}$ denote the current and preceding source latent frames, respectively, and $\mathbf{r}_{t,f}$ denotes the unrecovered latent discrepancy for latent frame $f$ at generation step $t$. $L_{t,f}$ is the frame-level semantic coding demand. $\rho_{t,f}$ is the bounded allocation score mapped from $L_{t,f}$. $M_{t,f}$ denotes the number of allocated codebook atoms.}

  \label{fig:bitrate_allocation}
\end{figure}
\subsection{Spatiotemporal Semantic Importance-Guided Bit Allocation}
\label{sec:temporal_allocation}

Given the candidate codebook at generation step $t$, uniformly assigning the same innovation number $M$ to every latent frame ignores the nonuniform coding demand across time. It may allocate unnecessary bits to stable frames while providing insufficient correction capacity for frames containing stronger structures, temporal changes, or unrecovered information.

As illustrated in Fig.~\ref{fig:bitrate_allocation}, our method summarizes these cues into a frame-level semantic coding demand. For latent frame $f$, it is defined as
\begin{equation}
\begin{aligned}
    L_{t,f}
    ={}&
    \alpha_s
    \mathcal{N}
    \left(
        \left\|
            \nabla_h\mathbf{y}_{0,f}
        \right\|_1
        +
        \left\|
            \nabla_w\mathbf{y}_{0,f}
        \right\|_1
    \right)
    \\
    &+
    \alpha_t
    \mathcal{N}
    \left(
        \left\|
            \mathbf{y}_{0,f}
            -
            \mathbf{y}_{0,f-1}
        \right\|_1
    \right)
    +
    \alpha_r
    \mathcal{N}
    \left(
        \left\|
            \mathbf{r}_{t,f}
        \right\|_1
    \right).
    \label{eq:frame_semantic_importance}
\end{aligned}
\end{equation}
The $\ell_1$ norm aggregates each cue over the channel and spatial dimensions, producing one scalar for each latent frame. The operator $\mathcal{N}(\cdot)$ normalizes each type of frame-level statistic over the latent frames at the same generation step. The coefficients $\alpha_s$, $\alpha_t$, and $\alpha_r$ control the relative contributions of spatial structure, temporal variation, and unrecovered discrepancy.
Unlike the position-wise map $\mathbf{W}_{t,f}$, which determines where the discrepancy should receive greater emphasis during atom selection, $L_{t,f}$ characterizes the overall coding demand of the entire latent frame.

We map $L_{t,f}$ to a bounded allocation score using
\begin{equation}
    \rho_{t,f}
    =
    \frac{
        L_{t,f}
    }{
        1+L_{t,f}
    }.
    \label{eq:semantic_allocation_score}
\end{equation}
This monotonic transformation preserves the relative ordering of frame-level demands while reducing the influence of extreme values. Given the minimum and maximum innovation budgets, the number of atoms assigned to latent frame $f$ is
\begin{equation}
    M_{t,f}
    =
    [M_{\min}
    +
    \left(
        M_{\max}
        -
        M_{\min}
    \right)
    \rho_{t,f}],
    \label{eq:temporal_bit_allocation}
\end{equation}
where $[\cdot]$ refers to the round operation.
Frames with larger coding demands receive more innovations, while fewer innovations are assigned to frames that can be recovered with weaker bitstream constraints. The allocated $M_{t,f}$ is used by the innovation-selection method in Section~\ref{sec:semantic_projection} to determine the selected codebook subset.  Since $M_{t,f}$ is derived from source-dependent information that is unavailable at the decoder, it is explicitly signaled in the bitstream for each generation-step and latent-frame pair. 

\begin{figure}[t]
  \centering
  \includegraphics[width=\linewidth]{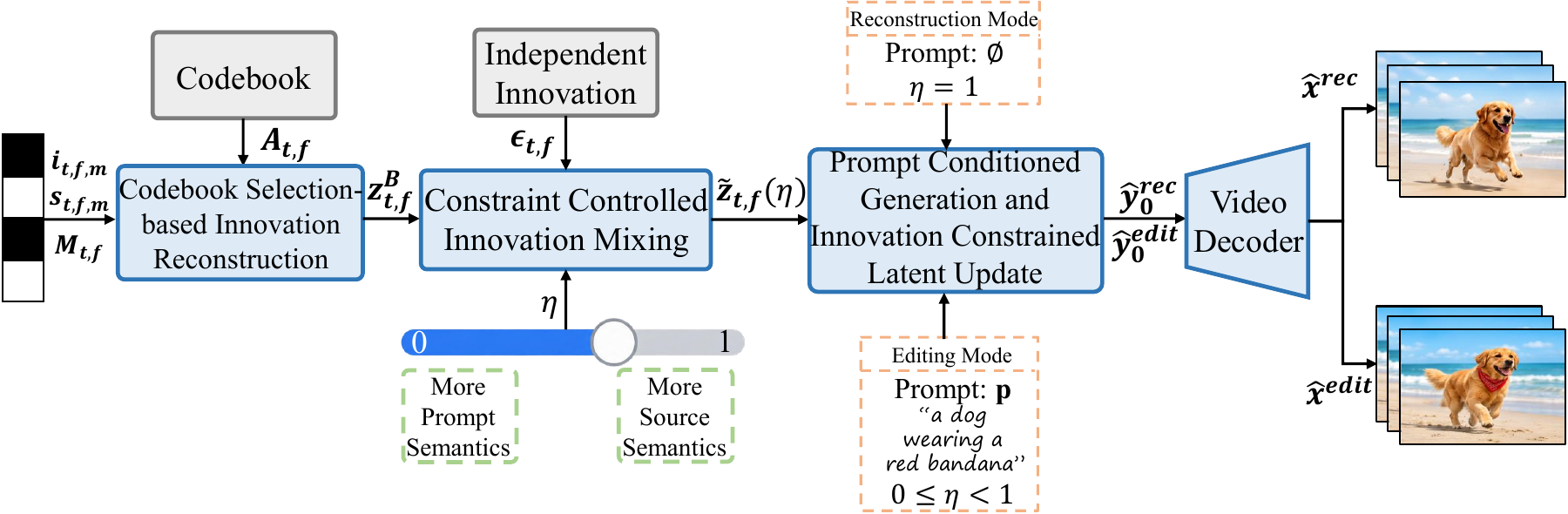}
  \caption{Illustration of the proposed spatiotemporal semantic importance-constrained unified reconstruction and editing. Here, $\mathbf{z}_{t,f}^{\mathcal{B}}$ is the reconstructed innovation. 
 $i_{t,f,m}$ and $s_{t,f,m}$ denote the index and sign of the $m$-th selected atom, respectively. The constraint strength $\eta$ controls the contribution of $\mathbf{z}_{t,f}^{\mathcal{B}}$, and $\boldsymbol{\epsilon}_{t,f}$ denotes an independent unit-variance innovation. Their combination produces the controlled innovation $\widetilde{\mathbf{z}}_{t,f}(\eta)$. Under the null-text condition and $\eta=1$, the reconstructed latent $\widehat{\mathbf{y}}_0^{\mathrm{rec}}$ and reconstructed video $\widehat{\mathbf{x}}^{\mathrm{rec}}$ are produced. Under a user-defined prompt and a relaxed constraint, the edited latent $\widehat{\mathbf{y}}_0^{\mathrm{edit}}$ and edited video $\widehat{\mathbf{x}}^{\mathrm{edit}}$ are produced.}

  \label{fig:unified_reconstruction_editing}
\end{figure}
\subsection{Spatiotemporal Semantic Importance-Constrained Unified Reconstruction and Editing}
\label{sec:prompt_editing}

Conventional video codecs decode a compressed bitstream into a reconstructed video. If users subsequently wish to modify the reconstructed content, a separate editing model typically needs to analyze the video again, for example by encoding it into another latent space to recover the structure and motion required for prompt-based editing. However, the transmitted innovations already constrain the spatiotemporal semantics of the source video. Re-extracting this information from the reconstructed pixels introduces redundant processing and additional computational complexity.

We therefore use the transmitted innovations directly as reusable semantic constraints for both reconstruction and editing. The main challenge is that the two tasks require different constraint strengths. Source reconstruction should fully preserve the recovered innovations, whereas prompt-based editing must relax their influence sufficiently for the new prompt to modify the content without losing the source structure and motion.

As illustrated in Fig.~\ref{fig:unified_reconstruction_editing}, the decoder first regenerates the shared codebook using the transmitted seed. It then retrieves the selected atoms according to the transmitted indices and the number of atoms $M_{t,f}$ and combines them with their corresponding signs following Eq.~\ref{eq:framewise_innovation}. Let $\mathbf{z}_{t,f}^{\mathcal{B}}$ denote the unit-variance innovation reconstructed from the transmitted indices and signs for latent frame $f$ at generation step $t$. We define the controlled innovation as
\begin{equation}
    \widetilde{\mathbf{z}}_{t,f}(\eta)
    =
    \eta
    \mathbf{z}_{t,f}^{\mathcal{B}}
    +
    \sqrt{1-\eta^2}\,
    \boldsymbol{\epsilon}_{t,f},
    \label{eq:semantic_constraint_mixing}
\end{equation}
where $\eta\in[0,1]$ controls the contribution of the bitstream constraint and $\boldsymbol{\epsilon}_{t,f}$ is an independent unit-variance innovation deterministically generated from the shared seed. Assuming independence between the two innovations, the coefficients preserve unit variance while continuously changing the contribution of the transmitted information. This avoids the variance reduction that would result from directly scaling $\mathbf{z}_{t,f}^{\mathcal{B}}$.

At each generation step, the frozen generator predicts the latent velocity under a text condition $\mathbf{c}$:
\begin{equation}
    \mathbf{u}_t(\mathbf{c})
    =
    G_\theta
    \left(
        \mathbf{y}_t,
        t,
        \mathbf{c}
    \right),
    \label{eq:conditioned_generator_velocity}
\end{equation}
where $\mathbf{c}=\varnothing$ for source reconstruction and $\mathbf{c}=\mathbf{p}$ for prompt-based editing. The corresponding SDE drift $\mathbf{f}_t(\mathbf{c})$ is obtained using the velocity-to-SDE conversion in Eqs.~\eqref{eq:velocity_to_score} and~\eqref{eq:sde_drift}. After stacking the controlled frame-wise innovations into $\widetilde{\mathbf{z}}_t(\eta)$, the latent is updated as
\begin{equation}
    \mathbf{y}_{t-\Delta t}
    =
    \mathbf{y}_t
    -
    \mathbf{f}_t(\mathbf{c})\Delta t
    +
    g_t\sqrt{\Delta t}\,
    \widetilde{\mathbf{z}}_t(\eta).
    \label{eq:constrained_stochastic_update}
\end{equation}

Starting from the same initial noisy latent $\mathbf{y}_T$, repeatedly applying Eq.~\eqref{eq:constrained_stochastic_update} with $\mathbf{c}=\varnothing$ and $\eta=1$ produces the reconstructed latent $\widehat{\mathbf{y}}_0^{rec}$. Applying the same updates with $\mathbf{c}=\mathbf{p}$ and $0\leq\eta<1$ produces the edited latent $\widehat{\mathbf{y}}_0^{edit}$. The corresponding output videos are
\begin{equation}
\begin{aligned}
    \widehat{\mathbf{x}}^{rec}
    &=
    \mathcal{D}
    \left(
        \widehat{\mathbf{y}}_0^{rec}
    \right),
    \\
    \widehat{\mathbf{x}}^{edit}
    &=
    \mathcal{D}
    \left(
        \widehat{\mathbf{y}}_0^{edit}
    \right),
\end{aligned}
\label{eq:unified_reconstruction_editing}
\end{equation}
where $\mathcal{D}$ denotes the frozen video decoder and $\mathbf{p}$ is the user-defined editing prompt.

For reconstruction, $\eta=1$ removes the independent innovation and fully retains the innovations recovered from the bitstream. Together with the null-text condition, this guides the generation trajectory toward the source video. For editing, an intermediate $\eta$ balances the source semantics conveyed by the bitstream with the editing semantics introduced by the prompt. A larger $\eta$ preserves more source structure and motion, whereas a smaller $\eta$ allows the prompt to exert a stronger influence. Consequently, the same compressed bitstream supports both source reconstruction and prompt-based editing by changing only the text condition and the constraint strength.

\begin{figure}[t]
  \centering
   \includegraphics[width=\linewidth]{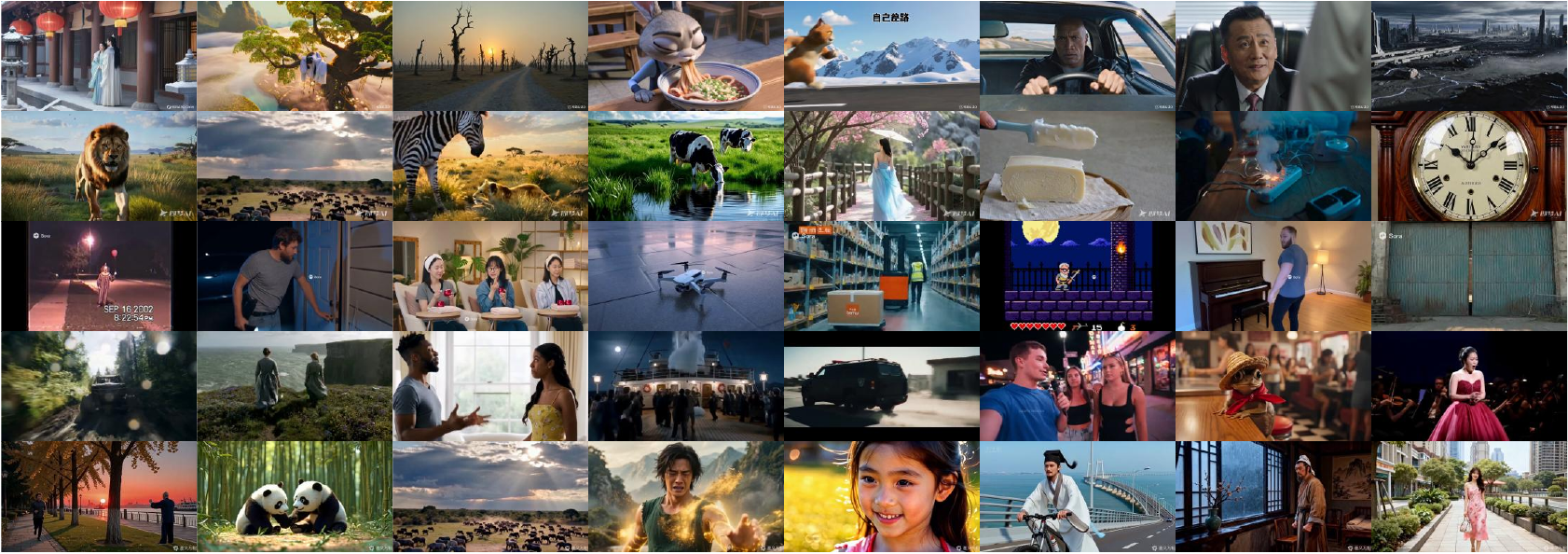}
      \caption{Representative test videos generated by Kling, Seedance, Sora, Veo, and Wan models. The test dataset contains 75 videos with diverse content, including human subjects, natural landscapes, objects, and animations.}
   \label{fig:test_sequences}
\end{figure}

\begin{figure*}[!htb]
  \centering
   \includegraphics[width=0.85\linewidth]{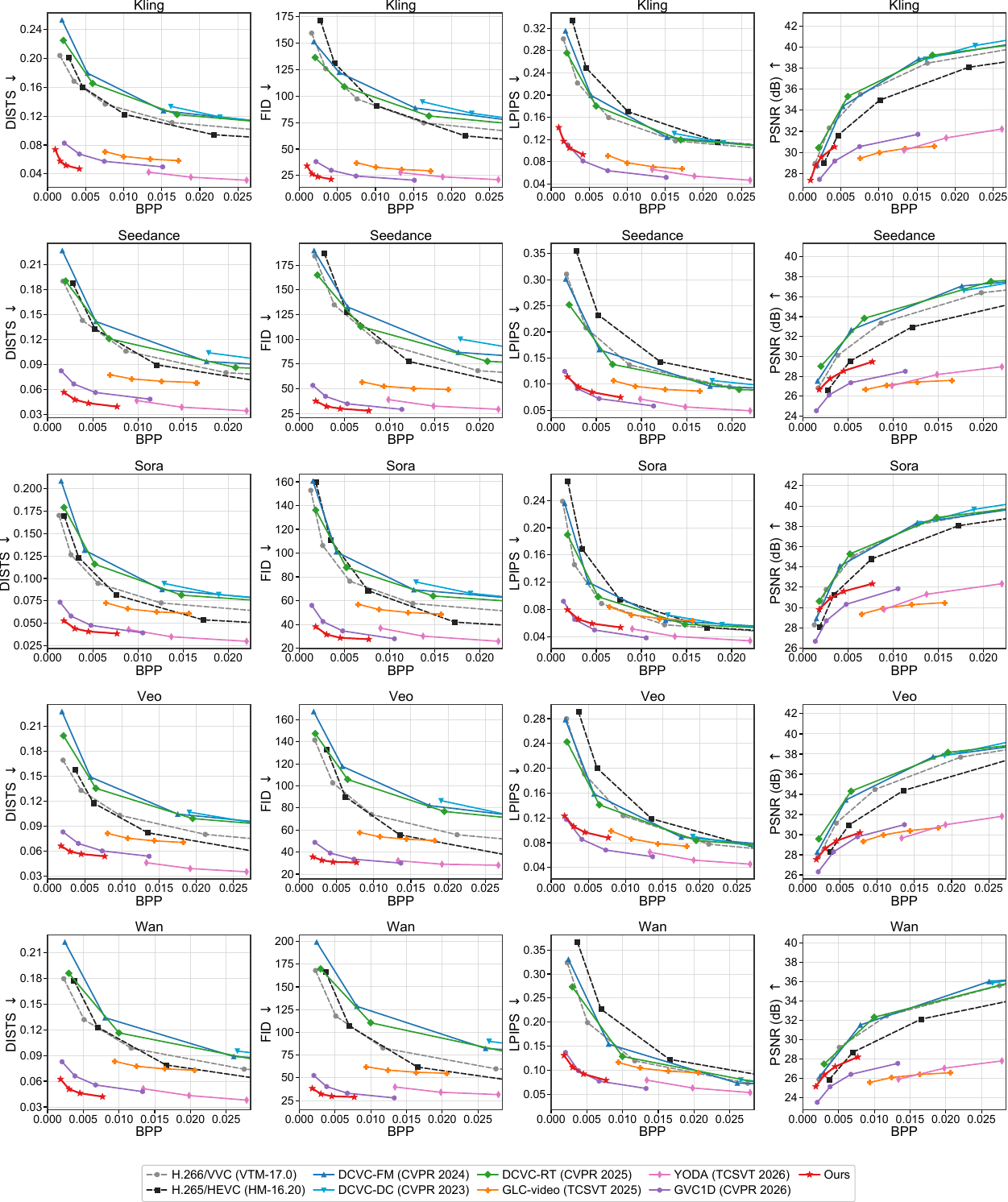}
      \caption{Quantitative comparison results on Kling, Seedance, Sora, Veo, and Wan test video sequences.}
   \label{fig:rd_results}
\end{figure*}
\section{Experiments}
\label{sec:experiments}
\subsection{Experimental Settings}
\label{sec:experimental_settings}

\subsubsection{Dataset}
We use the MVAD~\cite{hu2025mvad} dataset for evaluation, which contains AI-generated videos from five representative video generation models: Kling~\cite{team2025kling}, Seedance~\cite{seedance2026seedance}, Sora~\cite{liu2024sora}, Veo~\cite{wiedemer2025video}, and Wan~\cite{wan2025wan}. Each model contributes 15 test sequences, resulting in a total of 75 sequences, which is comparable in scale to test sets commonly used in neural video compression studies~\cite{sheng2025prediction,wang2024exploring,10159648,10129217}. The Kling sequences are evaluated at a resolution of $1920\times1080$, while the Seedance, Veo, Sora, and Wan sequences are evaluated at $1280\times720$. For each test sequence, we encode and evaluate the first 99 frames.
As illustrated in Fig.~\ref{fig:test_sequences}, the test sequences cover diverse generated content, including human subjects and activities, natural landscapes, object-centric scenes, and animated content. This diversity allows us to evaluate the compression performance of our method across different visual appearances, spatial structures, and motion characteristics.

\subsubsection{Implementation Details}
We implement the proposed framework using the frozen Wan2.1-T2V-14B model as the shared generative prior. Neither the video generator nor the video autoencoder is trained or fine-tuned. The generation step is set to 20. 
At each stochastic step and for each latent frame, the shared codebook contains $K=16{,}384$ independently sampled Gaussian atoms. The initial latent and all codebooks are deterministically generated using the shared seed $42$. For spatiotemporal semantic importance-guided innovation selection, the spatial, temporal, and discrepancy weights are set to $0.5$, $0.4$, and $0.3$, respectively. The temporal-consistency weight is set to $0.15$. 
For frame-adaptive bit allocation, the spatial, temporal, and discrepancy weights are set to $0.5$, $0.3$, and $0.2$, respectively.
Each $M_{t,f}$ is represented by a 16-bit unsigned integer. We vary the maximum innovation budget $M_{\max}$ to obtain different rate points.
All experiments are conducted on a single NVIDIA RTX Pro 6000 GPU.

\begin{table*}[t]
\centering
\caption{BD-rate (\%) of our codec relative to the state-of-the-art generative video codecs GVC1D and YODA on five AI-generated video classes.  A dash (---) indicates that BD-rate cannot be computed because the quality ranges do not sufficiently overlap. }
\label{tab:bdrate_anchor_comparison}
\setlength{\tabcolsep}{6.2pt}
\renewcommand{\arraystretch}{1.12}
\begin{tabular}{lrrrrrrrr}
\toprule[1.2pt]
\multirow{2}{*}{Class}
& \multicolumn{4}{c}{GVC1D Anchor}
& \multicolumn{4}{c}{YODA Anchor} \\
\cmidrule(lr){2-5}\cmidrule(lr){6-9}
& DISTS  & FID  & LPIPS & PSNR 
& DISTS  & FID  & LPIPS  & PSNR  \\
\midrule
Kling    & $-75.50$ & $-69.93$ & $  8.40$ & $-50.68$ & ---       & $-87.12$ & ---       & $-72.74$ \\
Seedance & $-70.29$ & $-57.77$ & $ 18.18$ & $-55.09$ & $-57.84$ & $-80.82$ & ---       & $-75.59$ \\
Sora     & $-51.27$ & $-52.63$ & $ 22.40$ & $-52.04$ & $-51.91$ & $-74.97$ & ---       & $-73.62$ \\
Veo      & $-59.74$ & $-64.28$ & $ 32.19$ & $-31.83$ & ---       & $-69.53$ & ---       & $-54.96$ \\
Wan      & $-66.52$ & $-58.31$ & $ -5.30$ & $-54.89$ & $-73.53$ & $-87.68$ & $-42.94$ & $-79.48$ \\
\midrule
\textbf{Average}
& $\mathbf{-64.66}$ & $\mathbf{-60.58}$ & $15.18$ & $\mathbf{-48.91}$
& $\mathbf{-61.09}$ & $\mathbf{-80.03}$ & $\mathbf{-42.94}$ & $\mathbf{-71.28}$ \\
\bottomrule[1.2pt]
\end{tabular}
\end{table*}

\begin{figure*}[t]
  \centering
   \includegraphics[width=0.8\linewidth]{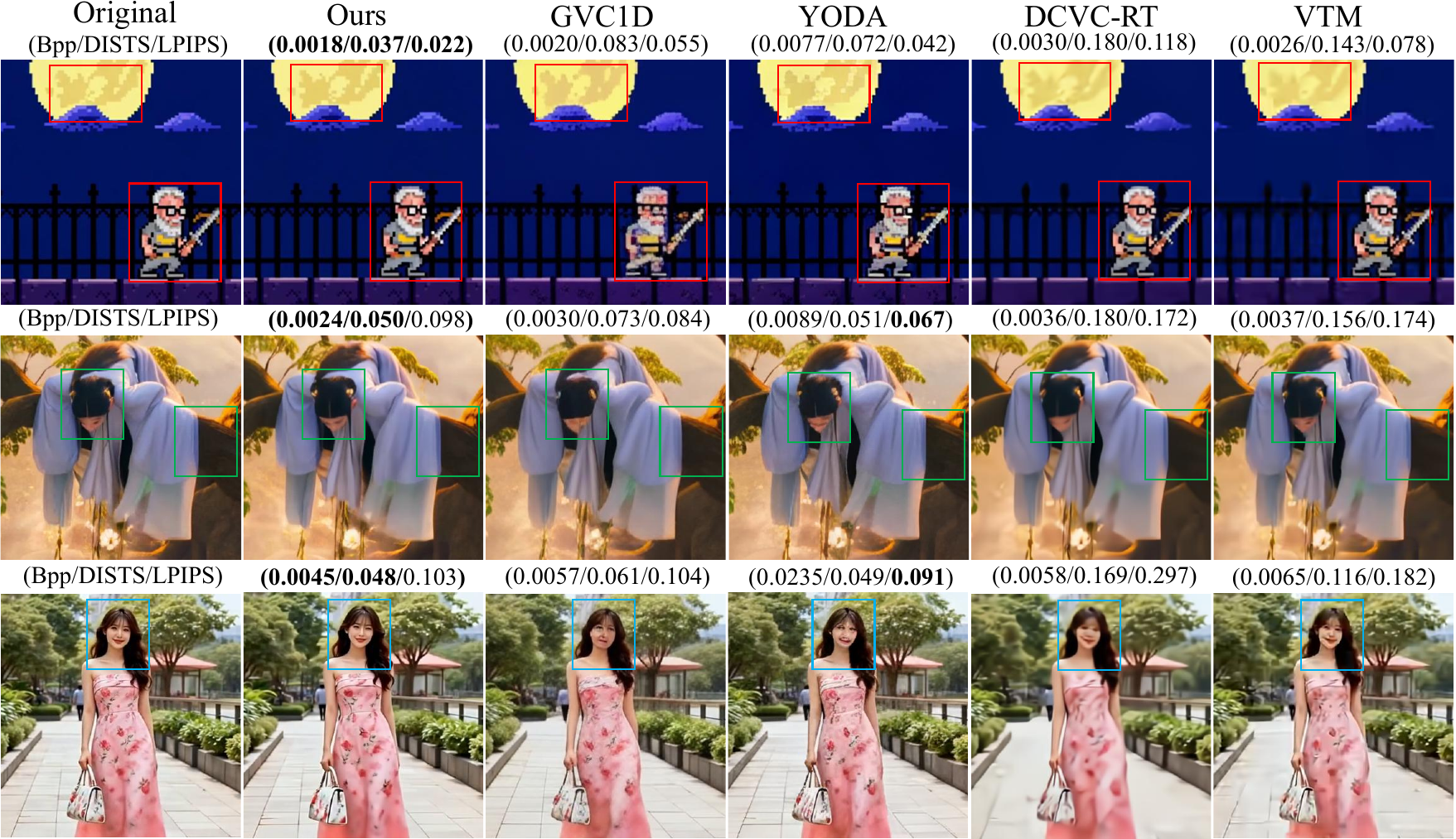}
      \caption{Qualitative reconstruction comparison results.}
   \label{fig:subjective_results}
\end{figure*}

\subsubsection{Compared Baselines}
We compare our codec with HM-16.20 and VTM-17.0, the reference software for H.265/HEVC~\cite{sullivan2012overview} and H.266/VVC~\cite{bross2021overview}, respectively. We also compare with the fidelity-oriented neural video codecs DCVC-DC~\cite{li2023neural}, DCVC-FM~\cite{li2024neural}, and DCVC-RT~\cite{jia2025towards}. For perceptual generative video compression, we evaluate GLC-video~\cite{qi2025generative}, GVC1D~\cite{zheng2026generative}, and YODA~\cite{li2026yoda}. 
We use the officially released implementations for all codecs. Following existing generative video coding studies, we measure reconstruction quality between the original and reconstructed frames in the RGB color space because the pretrained video generation prior operates in RGB space.
\begin{figure*}[htb]
  \centering
   \includegraphics[width=0.8\linewidth]{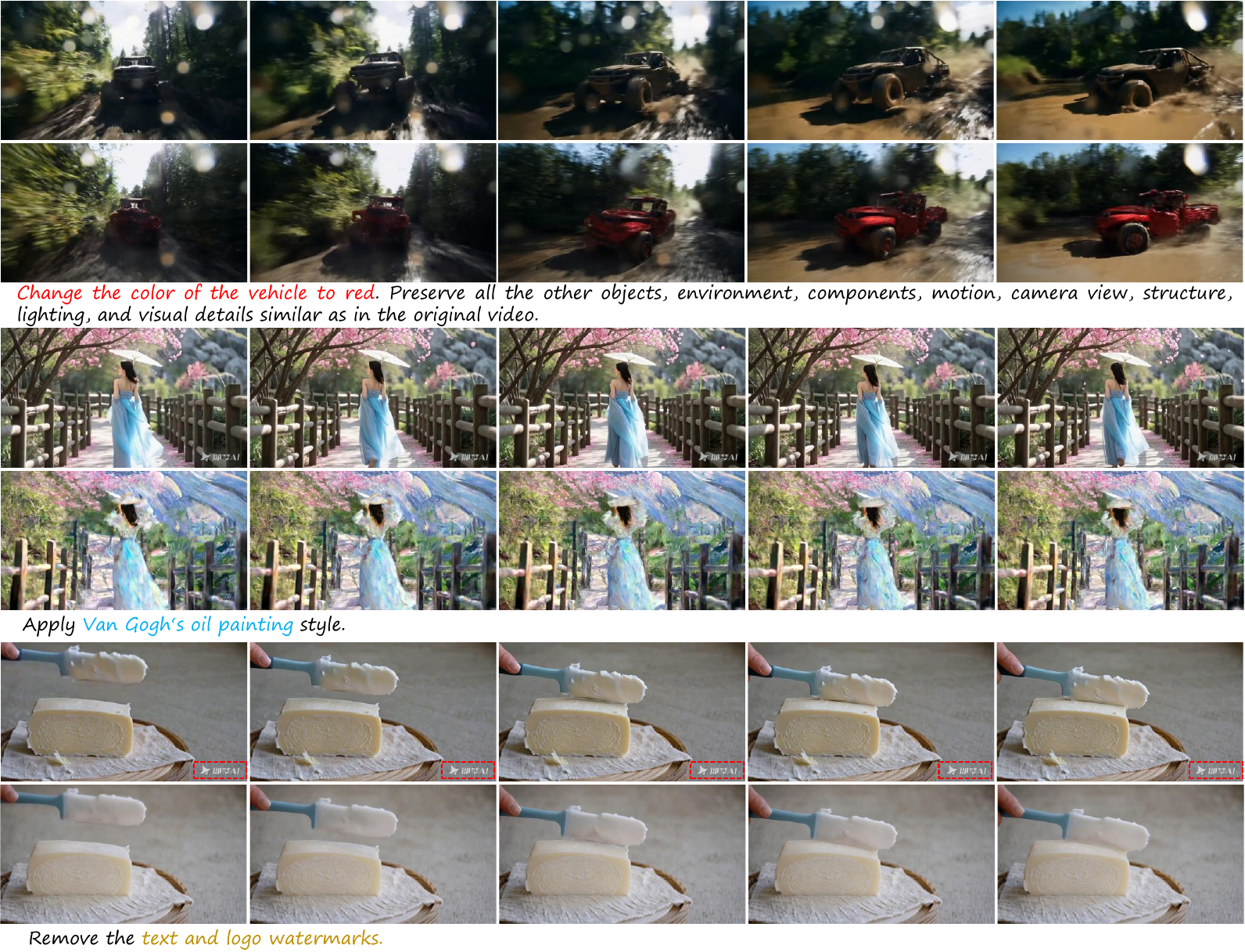}
      \caption{Reconstruction and prompt-controllable editing comparison results.}
   \label{fig:regeneration_results}
\end{figure*}
\subsubsection{Evaluation Metrics}
We measure the bitrate cost in bits per pixel (bpp) and evaluate reconstruction quality using DISTS~\cite{ding2020image}, FID~\cite{heusel2017gans}, LPIPS~\cite{zhang2018unreasonable}, and peak signal-to-noise ratio (PSNR). We compute LPIPS using AlexNet as the feature backbone. Following~\cite{qi2025generative}, we calculate FID on $256\times256$ image patches. PSNR primarily measures pixel-level fidelity, while LPIPS evaluates similarity in a learned feature space. However, PSNR and LPIPS metrics may not fully characterize perceptual quality under ultra-low-bitrate compression~\cite{zhang2025stablecodec}, where perceptually plausible reconstructions can differ from the source at the pixel or feature level. We therefore focus primarily on DISTS and FID, which provide complementary assessments of structural similarity and distribution-level perceptual quality, respectively. We use BD-rate (\%) to quantify the difference in compression performance between two codecs at equivalent reconstruction quality. A negative BD-rate indicates bitrate savings relative to the anchor codec, whereas a positive value indicates an increase in bitrate.

\subsection{Quantitative Reconstruction Comparison}
Figure~\ref{fig:rd_results} presents the rate--metric curves of different codecs on the five AI-generated video classes. Our framework consistently operates in a favorable low-bitrate region across different video generators. The advantages are particularly evident in DISTS and FID, demonstrating that our framework can preserve perceptual structures and improve distribution-level reconstruction quality using compact side information.
Table~\ref{tab:bdrate_anchor_comparison} reports the class-wise BD-rate results of our framework relative to the state-of-the-art generative video codecs GVC1D and YODA. Compared with GVC1D, our framework achieves average BD-rates of $-64.66\%$, $-60.58\%$, and $-48.91\%$ on DISTS, FID, and PSNR, respectively. Its average LPIPS BD-rate is 15.18\%, indicating that GVC1D retains an advantage under this particular metric because it optimizes LPIPS during training. With YODA as the anchor, our framework achieves average BD-rates of $-80.03\%$ on FID and $-71.28\%$ on PSNR over all five classes. Valid DISTS overlap is available for three classes, over which the average BD-rate is $-61.09\%$, while LPIPS has a valid overlapping range only for the Wan class, where the BD-rate is $-42.94\%$.

\subsection{Qualitative Reconstruction Comparison}
\label{sec:qualitative_results}
Figure~\ref{fig:subjective_results} presents qualitative reconstruction comparisons on Sora, Kling, and Wan test sequences. Our codec produces better subjective reconstruction quality across these diverse generation styles. Existing generative codecs such as GVC1D and YODA can obtain competitive perceptual metric scores (even achieve the best LPIPS performance for some sequences) and often reconstruct textures, repeated patterns, and background scenery reasonably well. However, they introduce noticeable semantic distortions. This failure is most evident in human face regions, where identity, facial geometry, and local structure can be changed even when the overall texture looks plausible. This semantic distortion appears across different types of AI-generated videos: the game-style Sora sequence, the animated Kling scene, and the realistic Wan scene all show visible face or identity degradation for existing generative codecs. In contrast, our codec better preserves both natural scene details and face semantics. This supports our central design principle: for AI-generated video compression, the bitstream should prioritize semantic structures rather than only producing perceptually plausible local textures.

\subsection{Prompt-Based Editing Comparison}
\label{sec:regeneration_results}

Figure~\ref{fig:regeneration_results} demonstrates that the same compressed bitstream can support both source reconstruction and prompt-based editing. In reconstruction mode, the transmitted side information fully constrains the generation process toward the source video. In editing mode, relaxing the constraint strength allows a user-defined prompt to modify the desired visual attributes while the bitstream preserves the source spatiotemporal semantics. For the Veo off-road vehicle sequence, the editing prompt changes the vehicle color to red while largely maintaining the scene layout, vehicle trajectory, and camera motion. For the Seedance landscape sequence, the prompt transforms the video into a Van Gogh-inspired oil-painting style while preserving its spatial composition and temporal dynamics. For the Seedance cheese-cutting sequence, the prompt removes the visible text and logo watermark while retaining the appearance of the cheese, the cutting action, and the surrounding scene. These examples demonstrate that our framework supports fine-grained attribute modification, global style transfer, and localized watermark removal from the same bitstream, without re-analyzing the reconstructed video or introducing a separate post-reconstruction editing model.

\begin{table}[t]
\centering
\caption{Ablation study of our proposed methods in terms of BD-rate (\%). }
\label{tab:ablation_components}
\setlength{\tabcolsep}{7.0pt}
\renewcommand{\arraystretch}{1.12}
\begin{tabular}{ccrr}
\toprule[1.2pt]
SSIGIS
& SSIGBA
& DISTS
& FID \\
\midrule
$\times$     & $\times$     &  0.00 &   0.00 \\
$\checkmark$ & $\times$     & --6.89 & --14.26 \\
$\checkmark$ & $\checkmark$ & --12.12 & --17.57 \\
\bottomrule[1.2pt]
\end{tabular}
\end{table}

\begin{figure}[t]
  \centering
   \includegraphics[width=0.9\linewidth]{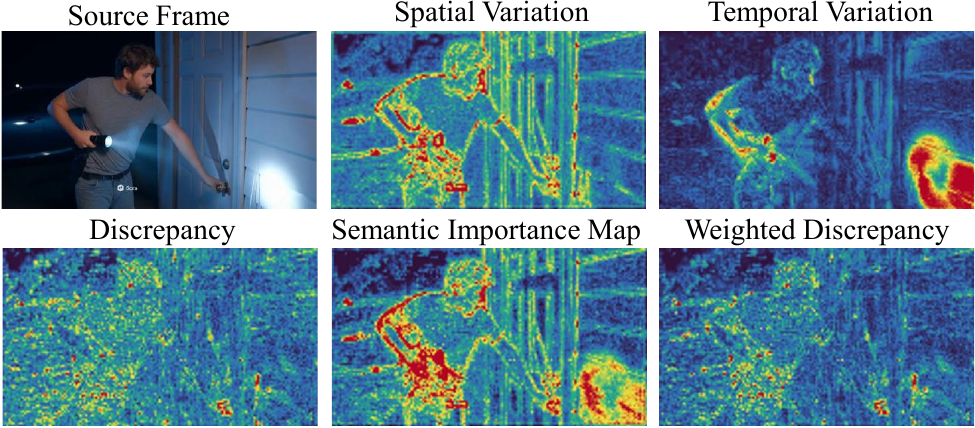}
      \caption{Visualization of the proposed spatiotemporal semantic importance-guided innovation selection. The top row shows the source video frame, spatial variation $\mathcal{N}(|\nabla_h\mathbf{y}_{0,f}|+|\nabla_w\mathbf{y}_{0,f}|)$, and temporal variation $\mathcal{N}(|\mathbf{y}_{0,f}-\mathbf{y}_{0,f-1}|)$. The bottom row shows the original discrepancy $|\mathbf{r}_{t,f}|$, semantic importance map $\mathbf{W}_{t,f}$, and weighted discrepancy $|\mathbf{r}_{t,f}^{\mathrm{sem}}|$.}
   \label{fig:ablation_semantic_selection}
\end{figure}

\section{Ablation Study}
\subsection{Effectiveness of Spatiotemporal Semantic Importance-Guided Innovation Selection}
We conduct an ablation study to verify the effectiveness of the proposed spatiotemporal semantic importance-guided innovation selection (SSIGIS). As reported in Table~\ref{tab:ablation_components}, introducing this method achieves an average bitrate saving of 6.89\% in terms of DISTS and 14.26\% in terms of FID, indicating that the same coding budget can be used more efficiently when innovation selection is guided by semantic importance.
To understand where this gain comes from, we visualize the source video frame, the spatial variation $\mathcal{N}(|\nabla_h\mathbf{y}_{0,f}|+|\nabla_w\mathbf{y}_{0,f}|)$, the temporal variation $\mathcal{N}(|\mathbf{y}_{0,f}-\mathbf{y}_{0,f-1}|)$, the original discrepancy $|\mathbf{r}_{t,f}|$, the semantic importance map $\mathbf{W}_{t,f}$, and the weighted discrepancy $|\mathbf{r}_{t,f}^{\mathrm{sem}}|=|\mathbf{W}_{t,f}\odot\mathbf{r}_{t,f}|$ in Fig.~\ref{fig:ablation_semantic_selection}. The example from the Sora class shows a person holding a flashlight and pushing open a door. The spatial variation identifies the person and object boundaries, while the temporal variation captures the arm motion and the displacement of the flashlight beam. In contrast, $|\mathbf{r}_{t,f}|$ alone cannot distinguish these semantic changes from stochastic errors in the dark background. By incorporating complementary spatial and temporal evidence, $\mathbf{W}_{t,f}$ assigns greater importance to discrepancies associated with structure and motion. Consequently, $|\mathbf{r}_{t,f}^{\mathrm{sem}}|$ makes these semantically meaningful discrepancies more dominant during codebook matching, explaining the observed coding gain.

\subsection{Effectiveness of Spatiotemporal Semantic Importance-Guided Bit Allocation}
\begin{figure}[t]
  \centering
   \includegraphics[width=0.9\linewidth]{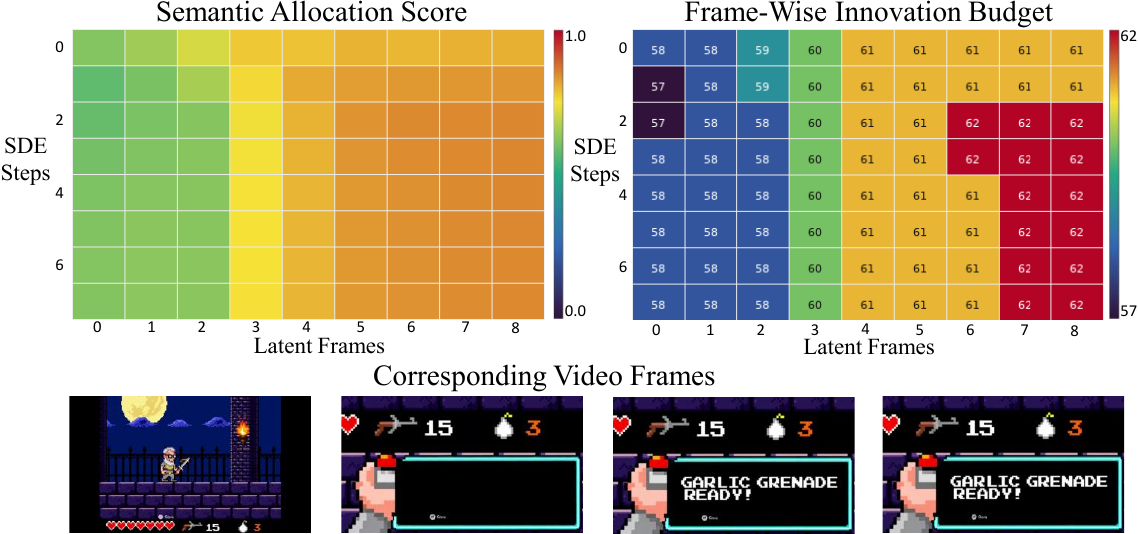}
      \caption{Visualization of the proposed spatiotemporal semantic importance-guided bit allocation. The first row shows the semantic allocation scores and corresponding frame-wise innovation budgets, while the second row presents representative video frames. Higher budgets are assigned to frames containing temporal scene transitions or complex spatial structures, such as text.}
   \label{fig:ablation_allocation}
\end{figure}

We evaluate the effectiveness of the proposed spatiotemporal semantic importance-guided bit allocation (SSIGBA) by comparing it with uniform allocation, where the same number of innovations is assigned to every latent frame. As reported in Table~\ref{tab:ablation_components}, introducing the proposed bit allocation method improves the BD-rate by 5.23\% in terms of DISTS and 3.31\% in terms of FID, demonstrating that distributing the coding budget according to frame-level semantic demand improves coding efficiency.
To further understand this improvement, we visualize the semantic allocation scores, the corresponding frame-wise innovation budgets, and video frames of a sequence from the Sora class in Fig.~\ref{fig:ablation_allocation}. The allocated budget varies consistently with changes in video content. In particular, the method assigns more innovations around temporal scene transitions, where the content changes substantially between adjacent frames. It also increases the budget for frames containing complex spatial structures, such as dense text and fine-grained details. These results show that the proposed method captures nonuniform coding demands across latent frames and concentrates coding resources on frames requiring stronger semantic preservation.

\begin{figure}[t]
  \centering
   \includegraphics[width=\linewidth]{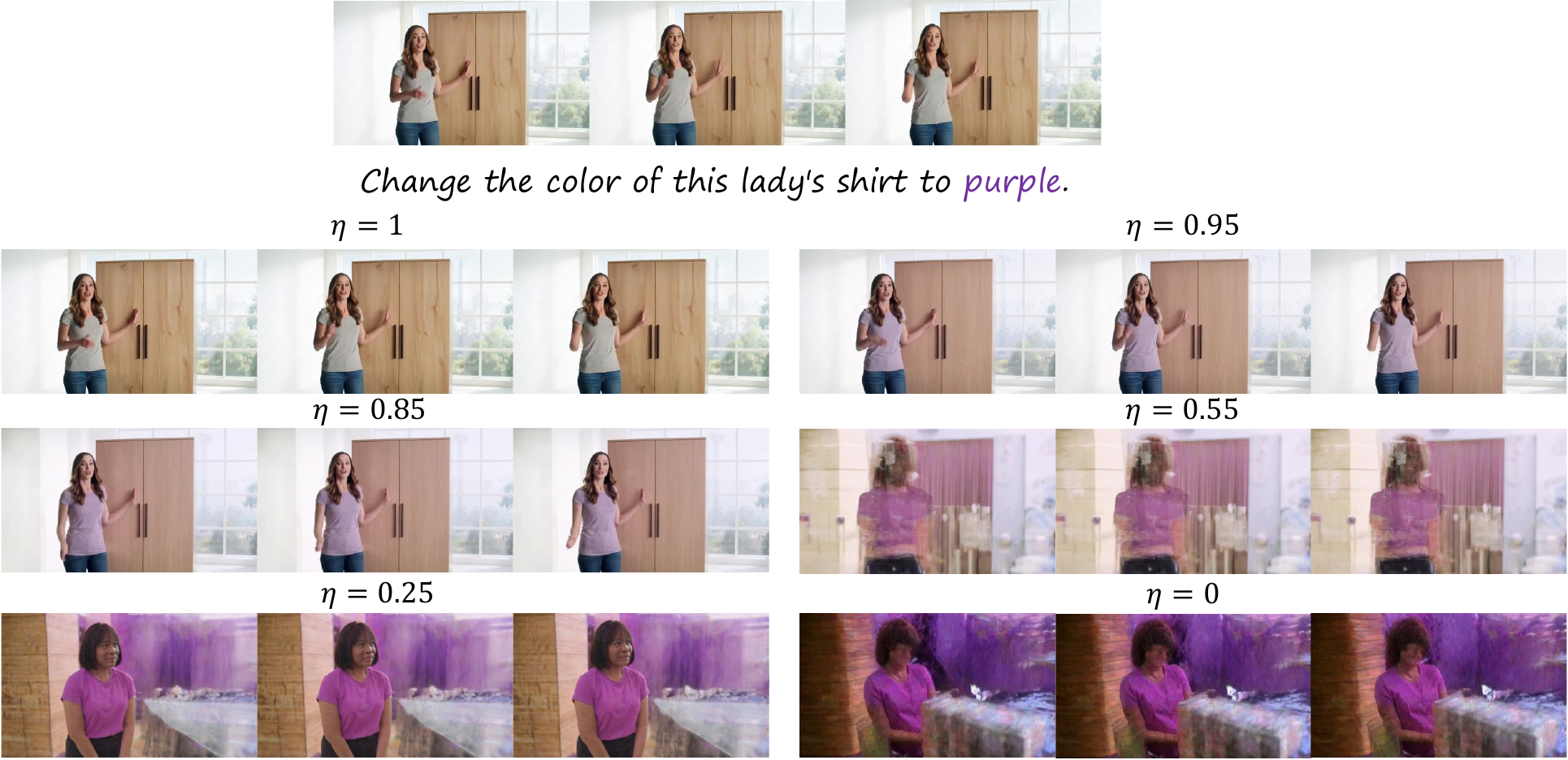}
\caption{Effect of the semantic constraint strength $\eta$. The editing prompt is ``Change the color of this lady's shirt to purple.'' As $\eta$ decreases, the influence of the transmitted source spatiotemporal semantics is progressively relaxed, producing increasingly stronger changes in appearance and motion.}
   \label{fig:constraint_strength}
\end{figure}
\subsection{Effectiveness of Spatiotemporal Semantic Importance-Constrained Unified Reconstruction and Editing}

We evaluate the effect of the semantic constraint strength $\eta$ using a video from the Veo class as an example. The editing prompt is ``Change the color of this lady's shirt to purple.'' As shown in Fig.~\ref{fig:constraint_strength}, we progressively decrease $\eta$ from $1$ to $0.95$, $0.85$, $0.55$, $0.25$, and $0$. When $\eta=1$, the transmitted innovations are fully retained, resulting in faithful reconstruction without prompt-induced modification. At $\eta=0.95$, the shirt is successfully changed to purple while the original action and scene structure remain largely unchanged. When $\eta$ decreases to $0.85$, the color modification becomes stronger, but noticeable changes in the woman's motion also begin to appear. Further reducing $\eta$ progressively weakens the source constraint and introduces increasingly substantial changes to the overall content and motion. At $\eta=0$, the transmitted innovations no longer constrain the generation process, and the output is primarily determined by the prompt and generative prior. These results demonstrate that $\eta$ provides continuous control over the trade-off between source spatiotemporal semantics and prompt-driven semantics, allowing the same bitstream to support faithful reconstruction and editing at different strengths.

\section{Conclusion and Future Work}
\label{sec:conclusion}
In this paper, we presented the first unified compression and editing framework specifically designed for AI-generated videos. Rather than treating an AI-generated video as a fixed natural signal whose every detail must be reproduced, our framework exploits its generative nature by using a frozen video generator to restore replaceable variations while transmitting compact side information that preserves source-specific spatiotemporal semantics. Within this framework, we developed three spatiotemporal semantic importance-guided techniques to determine what information should be transmitted, how the coding budget should be allocated across latent frames, and how the transmitted information should be used for reconstruction and editing. Experiments on videos produced by five representative generation models demonstrate favorable rate--perception performance over traditional, neural, and generative video codecs. They further show that the same compressed bitstream can support both faithful reconstruction and structure-preserving prompt-based editing.

These benefits currently come at the cost of relatively high computational complexity. For $720$p videos, the average encoding and decoding times are approximately $5.4$ and $4.6$ seconds per frame on a single NVIDIA RTX Pro 6000 GPU, respectively, mainly because both processes require iterative inference through the frozen video generator. This cost currently limits real-time deployment. However, recent advances in lightweight, on-device, and real-time video generation models~\cite{wu2025snapgen,kim2025device,karnewar2026neodragon,zhao2025real,feng2026streamdiffusionv2} provide promising opportunities to reduce generative inference costs. Future work will adapt the proposed framework to more efficient video generators and optimize its sampling and innovation-search processes for resource-constrained and real-time platforms. 
\bibliographystyle{ieeetr}
\bibliography{ref}
\end{document}